\documentclass[10pt,twocolumn,letterpaper]{article}

\usepackage{cvpr}              

\usepackage{graphicx}
\usepackage{amsmath}
\usepackage{amssymb}
\usepackage{booktabs}

\usepackage{amsfonts}       
\usepackage{nicefrac}       
\usepackage{microtype}      
\usepackage{xcolor}         
\usepackage{microtype}      
\usepackage{graphicx}
\usepackage{amsmath,amssymb}
\usepackage{mathrsfs}
\usepackage{multirow}
\usepackage{float}
\usepackage{setspace}
\usepackage{mathrsfs} 
\usepackage{enumitem}
\newfloat{figtab}{htb}{fgtb}
\makeatletter
  \newcommand\figcaption{\def\@captype{figure}\caption}
  \newcommand\tabcaption{\def\@captype{table}\caption}
\makeatother

\usepackage[pagebackref,breaklinks,colorlinks]{hyperref}

\usepackage[capitalize]{cleveref}
\crefname{section}{Sec.}{Secs.}
\Crefname{section}{Section}{Sections}
\Crefname{table}{Table}{Tables}
\crefname{table}{Tab.}{Tabs.}

\def\cvprPaperID{780} 
\def\confName{CVPR}
\def\confYear{2022}

\begin{document}

\title{Rethinking Bayesian Deep Learning Methods for\\ Semi-Supervised Volumetric Medical Image Segmentation}
 
\author{Jianfeng Wang\\
Department of Computer Science \\
University of Oxford \\ 
{\tt\small jianfeng.wang@cs.ox.ac.uk}
\and
Thomas Lukasiewicz\\
Department of Computer Science\\
University of Oxford \\
{\tt\small thomas.lukasiewicz@cs.ox.ac.uk}
}
\maketitle

\begin{abstract} 
   Recently, several Bayesian deep learning methods have been proposed for  semi-supervised medical image segmentation. Although they have achieved promising results on medical benchmarks, some problems are still existing. 
   Firstly, their overall architectures belong to the discriminative models, and hence, in the early stage of training, they only use labeled data for training, which might make them overfit to the labeled data. 
   Secondly, in fact, they are only partially based on Bayesian deep learning, as their overall architectures are not designed under the Bayesian framework. However, unifying the overall architecture under the Bayesian perspective can make the architecture have a rigorous theoretical basis, so that each part of the architecture can have a clear probabilistic interpretation. 
   Therefore, to solve the problems, we propose a new generative Bayesian deep learning (GBDL) architecture. GBDL belongs to the generative models, whose target is to estimate the joint distribution of input medical volumes and their corresponding labels. Estimating the joint distribution implicitly involves the distribution of data, so both labeled and unlabeled data can be utilized in the early stage of training, which alleviates the potential overfitting problem. Besides, GBDL is completely designed under the Bayesian framework, and thus we give its full Bayesian formulation, which lays a theoretical probabilistic foundation for our architecture.  Extensive experiments show that our GBDL outperforms previous state-of-the-art methods in terms of four commonly used evaluation indicators on three public medical datasets. 
\end{abstract}

\section{Introduction}
\label{sec:intro}

Deep neural networks are a powerful tool for visual learning, and they have outperformed previous methods that are based on hand-crafted features in many classical computer vision tasks. Due to the power of deep neural networks, researchers also try to use them to segment pathology regions from medical images. Training segmentation networks usually relies on large labeled medical image datasets, but they are scarce and difficult to  obtain, since not everyone is qualified to annotate medical data, and there is only a limited number of medical experts. Therefore, semi-supervised segmentation has become an important research direction in the area of medical computer vision. 

Several methods \cite{bai2017semi, zeng2021reciprocal, hang2020local, li2020shape,  huang20213d, bortsova2019semi, xie2020pairwise, luo2021semi, luo2021efficient, yu2019uncertainty, wang2020double, sedai2019uncertainty, shi2021inconsistency, wang2021tripled} have been proposed in recent years.\footnote{Note that we only consider previous methods that build their architectures with 3D CNNs, since our architecture is also based on 3D CNNs. In most cases, using 3D CNNs to process medical volumetric data performs better than using 2D CNNs, as the relationships among neighboring slices are preserved in 3D CNNs. Thus, comparing our method with 2D-CNN-based methods is unfair to them.} 
Among these works, Bayesian deep-learning-based methods \cite{yu2019uncertainty, wang2020double, sedai2019uncertainty, shi2021inconsistency, wang2021tripled} are closely related to this work, as they are based on Monte Carlo (MC) dropout \cite{gal2016dropout}, which is an approximation of Bayesian neural networks (BNNs). These methods are mainly built with the teacher-student architecture, which can be regarded as the discriminative model. Specifically, they initially build the model $P(Y|X)$ only from labeled data, and then apply the model to generate pseudo labels that are refined or rectified based on the epistemic uncertainty \cite{kendall2017uncertainties} provided by MC dropout for unlabeled ones. Then, the pseudo labels are further combined with unlabeled and labeled data to further train the overall architecture.

The works \cite{yu2019uncertainty, wang2020double, sedai2019uncertainty, shi2021inconsistency, wang2021tripled} mainly have two issues. Firstly, 
their models are built only with labeled data in the early stage of training and may overfit to them, since the quantity of labeled data is limited. Consequently, the new pseudo labels generated by the models might be inaccurate, which adversely impacts the subsequent training process. 
Secondly, their models are only partially based on Bayesian deep learning and are not designed under the Bayesian framework. Hence, they are
unable to give a Bayesian formulation with regard to their models, lacking a solid theoretical basis. 
As a result, some modules in their models are empirically designed, and the functions of those modules remain unclear.

In this work, we aim to fix these two problems by proposing a new generative model, named generative Bayesian deep learning (GBDL) architecture, as well as its corresponding full Bayesian formulation. 
Unlike  teacher-student-based architectures, GBDL estimates the joint probability distribution $P(X,Y)$ via both labeled and unlabeled data, based on which more reliable pseudo-labels can be generated, since the generative model is more faithful about the overall data distribution $P(X)$ and the intrinsic relationship between inputs and their corresponding labels (modeled by $P(X,Y)$), alleviating the overfitting problem.
Moreover, GBDL is entirely constructed under the Bayesian framework, and its related full Bayesian formulation lays a solid theoretical foundation, so that every part of GBDL has its corresponding probabilistic formulation and interpretation.

The main contributions of this paper are as follows:
\begin{itemize}[leftmargin=*, itemsep=0.5pt]
\item We propose a new generative Bayesian deep learning (GBDL) architecture. 
GBDL aims to capture the joint distribution of inputs and labels, which moderates the potential overfitting problem of the teacher-student architecture in previous works \cite{yu2019uncertainty, wang2020double, sedai2019uncertainty, shi2021inconsistency, wang2021tripled}. 

\item Compared with previous methods \cite{yu2019uncertainty, wang2020double, sedai2019uncertainty, shi2021inconsistency, wang2021tripled}, GBDL is completely designed under the Bayesian framework, and its full Bayesian formulation is also given here. Thus, it has a theoretical probabilistic foundation, and the choice of each module in GBDL and the construction of the loss functions are more mathematically grounded.

\item In extensive experiments,   GBDL outperforms previous methods with a healthy margin in terms of four evaluation indicators on three public medical datasets, illustrating that GBDL is a superior architecture for semi-supervised volumetric medical image segmentation.
\end{itemize}

The rest of this paper is organized as follows. In Section~\ref{sec:related_work}, we review related methods. Section~\ref{sec:methodology} presents our GBDL architecture and its  Bayesian formulation, followed by the experimental settings and results in Section~\ref{sec:experiment}. In Section~\ref{sec:conclusion}, we give a summary and 
an outlook on future research. 

The source code is  available at \href{https://github.com/Jianf-Wang/GBDL}{https://github.com/Jianf-Wang/GBDL}.

\section{Related Work}
\label{sec:related_work}

In this section, we briefly review related Bayesian and other deep learning methods for semi-supervised volumetric medical image segmentation in the past few years.

\subsection{Bayesian Deep Learning Methods} 
Previous Bayesian deep learning methods simply use MC dropout as a tool to detect when and where deep models might make false predictions \cite{yu2019uncertainty, wang2020double, sedai2019uncertainty, shi2021inconsistency, wang2021tripled}. Specifically, an uncertainty-aware self-ensembling model \cite{yu2019uncertainty} was proposed  based on the teacher-student model and MC dropout. In particular, a teacher model and a student model were built, where the latter learnt from the former by minimizing the segmentation loss on the labeled data and the consistency loss on all the input data. 
MC dropout was leveraged to filter out the unreliable predictions and to preserve the reliable ones given by the teacher model, and the refined predictions can help the student model learn from unlabeled data. Based on this uncertainty-aware teacher-student model, a double-uncertainty weighted method \cite{wang2020double} was proposed, which takes both the prediction and the feature uncertainty into consideration for refining the predictions of the teacher model during training. Sedai {\it et al.}~\cite{sedai2019uncertainty} used MC dropout for training the teacher model as well, but they designed a novel loss function to guide the student model by adaptively weighting regions with unreliable soft labels to improve the final segmentation performance, rather than simply removing the unreliable predictions. Shi {\it et al.}~\cite{shi2021inconsistency} improved the way of uncertainty estimation by designing two additional models, which are called object conservation model and object-radical model, respectively. Based on these two models, certain region masks and uncertain region masks can be obtained to improve pseudo labels and to prevent the possible error propagation during training.  Since the multi-task learning can boost the performance of segmentation, and it has not been considered in previous works, Wang {\it et al.}~\cite{wang2021tripled} combined it into their architecture, and imposed the uncertainty estimation on all tasks to get a tripled-uncertainty that is further used to guide the training of their student model. 

\subsection{Other Deep Learning Methods} 
Concerning other deep learning methods proposed recently, they are mainly classified as three categories: new training and learning strategies \cite{bai2017semi, zeng2021reciprocal}, shape-aware or structure-aware based methods \cite{hang2020local, li2020shape, huang20213d}, and consistency regularization based methods \cite{bortsova2019semi, xie2020pairwise, luo2021semi, luo2021efficient}. 

The new training and learning strategies aim  to gradually improve the quality of pseudo labels during training. For example, Bai {\it et al.}~\cite{bai2017semi} proposed an iterative training strategy, in which the network is first trained on labeled data, and then it predicts pseudo labels for unlabeled data, which are further combined with the labeled data to train the network again. During the iterative training process, a conditional random field (CRF) \cite{krahenbuhl2011efficient} was used to refine the pseudo labels. 
Zeng {\it et al.} \cite{zeng2021reciprocal} proposed a reciprocal learning strategy for their teacher-student architecture. The strategy contains a feedback mechanism for the teacher network via observing how pseudo labels would affect the student, which is omitted in previous teacher-student based models. 

The shape-aware or structure-aware based methods encourage the network to explore complex geometric information in medical images. For instance, 
Hang {\it et al.}~\cite{hang2020local} proposed a local and global structure-aware entropy-regularized mean teacher (LG-ER-MT) architecture for semi-supervised segmentation. It extracts local spatial structural information by calculating inter-voxel similarities within small volumes and global geometric structure information by utilizing weighted self-information. Li {\it et al.}~\cite{li2020shape} introduced a shape-aware semi-supervised segmentation strategy that integrates a more flexible geometric representation into the network for boosting the performance. Huang {\it et al.} \cite{huang20213d} designed a 3D Graph Shape-aware Self-ensembling Network (3D Graph-$S^2$Net), which is composed by a multi-task learning network and a graph-based module. The former performs the semantic segmentation task and predicts the signed distance map that encodes
rich features of object shape and surface, while the latter explores co-occurrence relations and diffuse information between these two tasks.  

The consistency-regularization-based methods try to add different consistency constraints for unlabeled data. In particular, 
Bortsova {\it et al.}~\cite{bortsova2019semi} proposed a model that is implemented by a Siamese architecture with two identical branches to receive differently transformed versions of the same images. Its target is to learn the consistency under different transformations of the same input. 
Luo {\it et al.} \cite{luo2021semi} designed a novel dual-task-consistency
model, whose basic idea is to encourage consistent predictions of the same input under different tasks. Xie {\it et al.}~\cite{xie2020pairwise} proposed a pairwise relation-based semi-supervised (PRS2) model for gland segmentation on histology tissue images. In this model, a supervised segmentation network (S-Net) and an unsupervised pairwise relation network (PR-Net) are built. The PR-Net learns both semantic consistency and image representations from each pair of images in the feature space for improving the segmentation performance of S-Net. Luo {\it et al.} \cite{luo2021efficient} proposed a 
pyramid (i.e., multi-scale) architecture that encourages the predictions of an unlabeled input at multiple scales to be consistent, which serves as a regularization for unlabeled data.

\section{Methodology}
\label{sec:methodology} 

In this section, we start from giving the Bayesian formulation of our GBDL architecture. Thereafter, the architecture and related loss functions are introduced in detail.

\subsection{Bayesian Formulation}

\paragraph{Learning Procedure.} The objective of semi-supervised volumetric medical image segmentation is to learn a segmentation network, whose weights are denoted by $W$, with partially labeled data, and the Bayesian treatment of this target is to learn the posterior distribution, namely,  $P(W|X, Y_{L})$.  
Now, we use $X$ to denote the input volumes, which contains labeled input volumes ($X_L$) and unlabeled input volumes ($X_U$), i.e., $X=\{ X_L, X_U \}$, and we use $Y=\{ Y_L, Y_U \}$ to denote the ground-truth labels with respect to $X$. 
Each label in $Y$ is a voxel-wise segmentation label map that has the same shape as its corresponding input volume. Note that $Y_U$ represents the ground-truth labels of $X_U$, which are not observed in the training data. 

The whole learning procedure of $P(W|X, Y_{L})$ can be written as:
\begin{equation}
\begin{aligned}
\label{eqn:overall}
& P(W|X, Y_{L}) \\
& = \iiint P(W|X, Y)P(X, Y|Z)P(Z|X, Y_{L})dZdXdY,
\end{aligned}
\end{equation}
where $Z$ is the latent representation that governs the  joint distribution of $X$ and $Y$, denoted by $P(X,Y|Z)$. To estimate the joint distribution, one should learn $Z$ from $X$ and $Y_{L}$. Concerning that  Eq.~(\ref{eqn:overall}) is intractable, we take the MC approximation of it as follows:
\begin{equation}
\label{eqn:monte_appro}
\begin{aligned}
 P(W|X, Y_{L}) = \frac{1}{MN}\sum\nolimits_{i=0}^{N-1}\sum\nolimits_{j=0}^{M-1}P(W|X_{(i,j)}, Y_{(i,j)}).
\end{aligned}
\end{equation}
In this approximation, $M$ latent representations $Z$ are drawn from $P(Z|X, Y_{L})$. Then, $N$ pairs of input volumes and labels are obtained from $P(X, Y|Z)$, which are further used to get the posterior distribution. Thus, to generate $X$ and $Y$, one should obtain the joint distribution $P(X, Y)$, i.e., estimating the distribution of $Z$. Overall, the learning procedure contains two steps: 
\begin{itemize}[leftmargin=*, itemsep=0.5pt]
\item Learning the distribution of $Z$ that governs the joint distribution $P(X, Y)$. 

\item Learning the posterior distribution $P(W|X, Y)$ based on $X$ and $Y$ sampled from $P(X, Y)$.
\end{itemize}

\paragraph{Inference Procedure.} The inference procedure can be formulated as follows:
\begin{equation}
\begin{aligned}
\label{eqn:testing}
&P(Y_{pred}|X_{test}, X, Y_{L})\\
&  =  \int P(Y_{pred}|X_{test}, W)P(W|X, Y_{L}) dW,
\end{aligned}
\end{equation}
where $X_{test}$ and $Y_{pred}$ denote the test input and the predicted result, respectively. The posterior distribution $P(W|X, Y_{L})$ is learnt based on $X$ and $Y$ sampled from $P(X, Y)$, and thus the posterior can also be replaced by $P(W|X, Y)$. Since Eq.~(\ref{eqn:testing}) is intractable as well, its MC approximation can be written as:
\begin{equation}
\label{eqn:testing_monte}
\begin{aligned}
&P(Y_{pred}|X_{test}, X, Y_{L}) = \frac{1}{T}\sum\nolimits_{i=0}^{T-1}P(Y_{pred}|X_{test}, W_i),
\end{aligned}
\end{equation}
where $T$ models are drawn from $P(W|X, Y)$, which is usually implemented via MC dropout \cite{gal2016dropout} with $T$ times feedforward passes. For each model ($W_i$) sampled from $P(W|X, Y)$, a prediction result can be obtained, and the final prediction of $X_{test}$ can be calculated by averaging the $T$ results. In addition, 
we get the epistemic uncertainty by calculating the entropy of these predictions.

\begin{figure*}[t]
\centering
\includegraphics[width=0.98\linewidth]{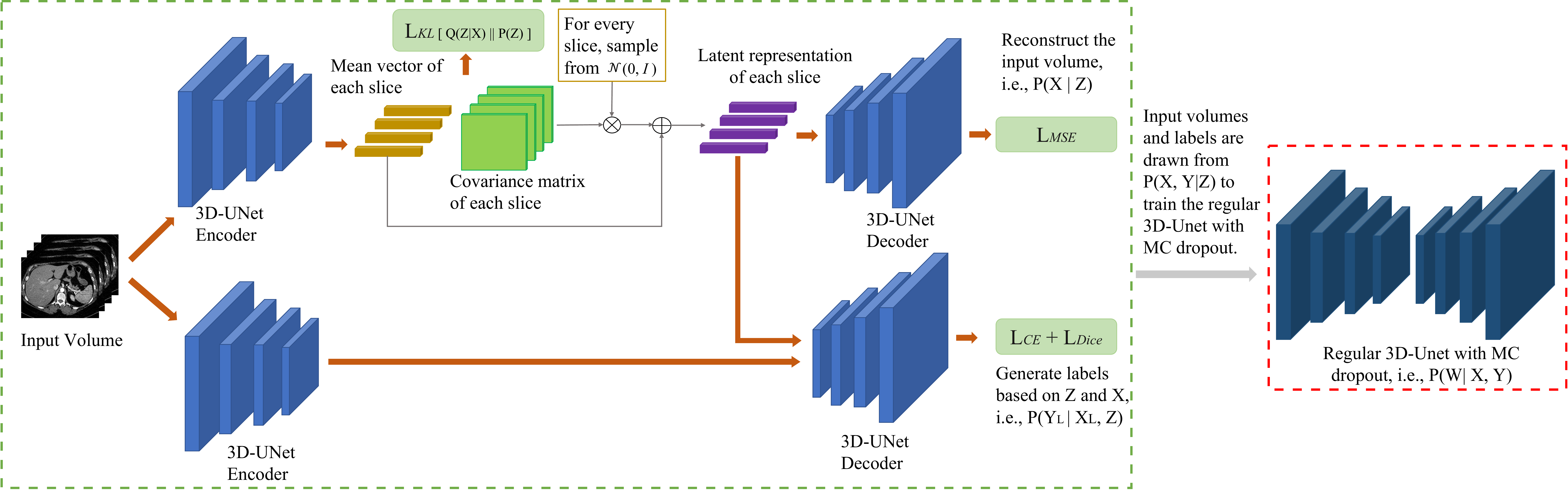} 
\caption{GBDL for semi-supervised volumetric medical image segmentation, including a latent representation learning (LRL) architecture (in the green dotted box) and a regular 3D-UNet with MC dropout (in the red dotted box). Only the regular 3D-UNet with MC dropout is used during testing. For simplicity, the shortcut connections between the paired 3D-UNet encoder and decoder are omitted.}
\vspace{-2ex}
\label{fig:framework}
\end{figure*} 

\subsection{GBDL and Loss Functions} 
\label{sec:gbdl} 

Under the guidance of the given Bayesian formulation, we start to introduce the details of GBDL and loss functions. 
Concerning the two steps mentioned in the learning procedure above, GBDL has a latent representation learning (LRL) architecture and a regular 3D-UNet with MC dropout, which are shown in the green dotted box and the red dotted box in Figure~\ref{fig:framework}, respectively. LRL is designed to learn the distribution of $Z$ and to capture the joint distribution $P(X, Y)$, while the regular 3D-UNet with MC dropout is the Bayesian deep model used for parameterizing  the posterior distribution $P(W|X, Y)$. 

As for LRL, we first assume a variational probability distribution $Q(Z)$ to represent the distribution of the latent representation $Z$. 
In this work, we follow the conditional variational auto-encoder (cVAE) \cite{sohn2015learning} to use the input to modulate $Z$, and thus, $Q(Z)$ can be rewritten as $Q(Z|X)$. 
Note that this work focuses on using 3D CNNs to process volumetric medical data, and therefore, 
each input volume contains several slices, $Q(Z|X)$ is actually the joint distribution of all these slices, i.e., $Q(Z|X) = q(z_0, z_1, z_2, ..., z_{n-1}|x_0, x_1, x_2, ..., x_{n-1})$, where $n$ is the number of slices in an input volume, $x_i$ and $z_i$ are a slice and its latent representation,  respectively. To obtain the latent representation of each slice $z_i$, a 3D-UNet encoder\footnote{In this paper, all 3D upsampling and downsampling layers used in the 3D-UNet encoder and the 3D-UNet decoder only perform on the spatial size, and therefore the depth of the input volumes keeps unchanged, and we can obtain the latent representation of each slice. More details about the network configuration are given in the supplementary material.} is used in LRL. As the downsampling layers of the 3D-UNet encoder do not perform on the depth, and the 3D convolutional layers used in the encoder are of the kernel size $3$, the padding size $1$, and the stride $1$ (shown in the supplementary material), each $z_i$ is actually conditioned on $n_{rf}$ slices, where $n_{rf}$ is the total receptive field along the depth dimension, which is determined by the number of 3D convolutional layers in the encoder.  
For ease of calculation, we assume that the distribution of the latent representation of each slice follows a multivariate Gaussian distribution and that the latent representations of slices are independent from each other. 
It is worth noting that such independence assumption on the latent representations of slices is reasonable, as in this way, the latent representation of each slice just rests on $n_{rf}$ slices, which is consistent with the fact that each slice is closely relevant to its neighboring slices and distant slices may have no contribution to the slice. 
Therefore, based on the independence assumption, $Q(Z|X)$ can be decomposed into $\prod_{i=0}^{n-1} q(z_i| x_{(n_{rf})})$, where $x_{(n_{rf})}$ denotes the input slices that contribute to the $z_{i}$. 
After obtaining the latent representation $Z$, we can write the evidence lower bound (ELBO) as follows (with proof in the supplementary material):
\begin{equation}
\begin{aligned}
\label{eqn:elbo}
 &logP(X,Y) \geq  \\
 &\mathbb{E}_{Q}[logP(Y| X, Z) + logP(X|Z)] - \mathbb{E}_{Q}[log (\frac{Q(Z|X)} {P(Z)})],
\end{aligned}
\end{equation}
where $\mathbb{E}_{Q}$ denotes the expectation over $Q(Z|X)$.  Thus, the learning objective is to maximize the ELBO (Eq.~(\ref{eqn:elbo})), which is achieved by maximizing $ \mathbb{E}_{Q}[logP(Y| X, Z) + logP(X|Z)]$ and minimizing~$\mathbb{E}_{Q}[log({Q(Z|X)}/{P(Z)})]$.

Firstly, to maximize $P(X|Z)$, another 3D-UNet decoder is used to take $z_i$ as input and the mean square error ($L_{MSE}$) is used as the loss function, which aims to reconstruct each slice of the input volume. Secondly, to maximize the probability $P(Y| X, Z)$, another branch can be built, which contains a 3D-UNet encoder and a 3D-UNet decoder as well. Since only parts of data have labels, $P(Y| X, Z)$ can also be rewritten as $P(Y_L| X_L, Z)$. 
In this branch, the 3D-UNet encoder receives $X_L$ to extract features that are further combined with their corresponding $Z$ to be input to the 3D-UNet decoder. Then, $Y_L$ is leveraged to calculate the dice loss $L_{Dice}$ \cite{milletari2016v} and cross-entropy loss $L_{CE}$ with $X_L$ to maximize the probability $P(Y_L| X_L, Z)$. 
Thirdly, minimizing $\mathbb{E}_{Q}[log({Q(Z|X)}/{P(Z)})]$ is to use $Q(Z|X)$ to approximate the prior distribution of $Z$, and in most cases, assuming that $Q(Z|X)$ and $P(Z)$ follow the same probability distribution can make the computation convenient.
The following theorem (with proof in the supplementary material) is useful for constructing the distribution of $Q(Z|X)$.

\smallskip
{\bf Theorem 1.} \emph{The product of any number $n\ge 1$ of multivariate Gaussian probability density functions with precision $\Lambda_i$ and mean $\mu_i$ results in an unnormalized Gaussian curve with precision $\Lambda_{*}$ $=$ $\sum_{i=0}^{n-1}\Lambda_i$ and mean $\mu_{*}\,=$ $\Lambda_{*}^{-1} (\sum_{i=0}^{n-1}\Lambda_i\mu_i)$.}
\smallskip

Although an unnormalized Gaussian curve (denoted $f(x)$) is obtained by multiplying several multivariate Gaussian PDFs, a Gaussian distribution can still be obtained by simply normalizing $f(x)$ with $\int f(x)dx$ that does not contain $x$. Therefore, based on {Theorem 1}, we can suppose that $Q(Z|X)$ follows a multivariate Gaussian distribution. In addition, we also assume that the prior distribution $P(Z)$ follows a multivariate standard normal distribution. To minimize $\mathbb{E}_{Q}[log ({Q(Z|X)}/{P(Z)})]$, we propose a new loss function  as follows (with proof in the supplementary material). 

\smallskip
{\bf Corollary 1.} \emph{Minimizing $\mathbb{E}_{Q}[log({Q(Z|X)}/{P(Z)})]$ is equivalent to minimizing}
\begin{equation}
\begin{aligned} 
\label{eqn:kl1}
& L_{KL[Q(Z|X)||P(Z)]}  = \\ 
& 1/2 \cdot (
-log\ det [(\sum\nolimits_{i=0}^{n-1}\Lambda_i)^{-1}] +  tr[(\sum\nolimits_{i=0}^{n-1}\Lambda_i)^{-1}] \ +  \\
 & (\sum\nolimits_{i=0}^{n-1}\Lambda_i\mu_i)^T (\sum\nolimits_{i=0}^{n-1}\Lambda_i)^{-2} (\sum\nolimits_{i=0}^{n-1}\Lambda_i\mu_i)
- D), 
\end{aligned}
\end{equation}
\emph{where $n$ denotes the number of slices, and $D$ denotes the number of dimensions of mean vectors.}
\smallskip

To summarize, maximizing ELBO (Eq.~(\ref{eqn:elbo})) is equivalent to optimizing LRL and minimizing the loss function:
\begin{equation}
\begin{aligned}
\label{eqn:ELBO_loss}
& L_{ELBO} =  \\
& \lambda_1 L_{CE} + \lambda_2 L_{Dice} + \lambda_3  L_{MSE} + \lambda_4  L_{KL[Q(Z|X)||P(Z)]},
\end{aligned}
\end{equation}
where $\lambda_1$, $\lambda_2$, $\lambda_3$, and $\lambda_4$  are coefficients. 
{Significantly}, cVAE \cite{sohn2015learning} has some similar properties when compared with our LRL architecture, and it is important to emphasize the differences between cVAE-based models and our LRL, which mainly contain three aspects:

\begin{itemize}[leftmargin=*, itemsep=0.5pt]
\item Different ELBO: cVAE is mainly used for generation tasks, in which the data distribution $P(X)$ is considered, and labels $Y$ are not involved into the ELBO of cVAE. In contrast, our LRL concerns  the joint distribution of $X$ and $Y$, leading to a different ELBO (Eq.~(\ref{eqn:elbo})).

\item Independent Components Assumption: 
cVAE assumes that the components in the latent representation of an input image are independent from each other. As a result, the encoder of cVAE only gives a mean vector and a variance vector for a latent representation\footnote{In practice, the encoder of cVAE gives log-variance vectors instead of variance vectors.}. However, our LRL does not follow this assumption for each slice, and the encoder gives a mean vector and a covariance matrix for the latent representation of each slice\footnote{Although the independent components assumption is widely used, it could be too strong and may degrade the performance.}.

\item Joint Variational Distribution: cVAE is mainly used for 2D inputs. In contrast, our LRL is specifically designed for 3D-CNN-based architectures for processing volumetric inputs. Therefore, since each input volume contains several slices, the variational distribution becomes a joint distribution over their latent representations, resulting in a different formulation of Kullback–Leibler (KL) divergence between the joint variational distribution and the prior distribution (Eq.~(\ref{eqn:kl1})).
\end{itemize}

Once LRL is trained well, the joint distribution of $X$ and $Y$ will be captured, and 
the second step is to learn the posterior distribution $P(W|X, Y)$. Therefore, input volumes and corresponding labels are generated from the fixed LRL to train the regular 3D-UNet with MC dropout with two loss functions that are widely used in medical image segmentation, namely,  the dice loss $L_{Dice}$ and the cross-entropy loss $L_{CE}$. In this paper, we use $L_{Seg}$ to denote their summation: 
\begin{equation}
\label{eqn:segment_loss}
L_{Seg} =  \beta_1 L_{CE} + \beta_2 L_{Dice},
\end{equation}
where $\beta_1$ and $\beta_2$ are the coefficient of the two loss terms in $L_{Seg}$. In the real implementation, the new generated pseudo labels will be combined with the unlabeled data and labeled data to train the regular 3D-UNet with MC dropout. According to  Eq.~(\ref{eqn:testing_monte}), only the the regular 3D-UNet with MC dropout is used in the test phase. 
In addition, as we do a full Bayesian inference during testing, it is convenient to obtain the voxel-wise epistemic uncertainty for each predicted result by calculating the entropy ($-\sum_{c=0}^{C-1}p_clog_2p_c$) voxel-by-voxel, where $C$ is the number of classes.

\begin{table*}
\centering
\resizebox{0.95\textwidth}{!}{
\begin{tabular}{@{}c|cc|cccc|cc|cccc@{}}
 \toprule[1pt]
 \multirow{3}{*}{} &  \multicolumn{6}{c|}{KiTS19} & \multicolumn{6}{c}{Atrial Segmentation Challenge} \\
  \cline{2-13}
  &  \multicolumn{2}{c|}{Scans used} &\multicolumn{4}{c|}{Metrics} & \multicolumn{2}{c|}{Scans used} &\multicolumn{4}{c}{Metrics}\\
 \cline{2-13}
 & Labeled & Unlabeled & Dice $\uparrow$ & Jaccard  $\uparrow$ & 95HD $\downarrow$ & ASD $\downarrow$ & Labeled & Unlabeled & Dice $\uparrow$ & Jaccard  $\uparrow$ & 95HD $\downarrow$ & ASD $\downarrow$ \\
 \hline
\multirow{3}{*}{\shortstack{3D-UNet \\  w/ MC dropout}} & 160 &  0 &  0.940 {\scriptsize($\pm$0.004)}  & 0.888 {\scriptsize($\pm$0.005)}   & 3.66 {\scriptsize($\pm$0.26)}  &  0.78 {\scriptsize($\pm$0.11)} & 80 &   0  & 0.915 {\scriptsize($\pm$0.006)}  & 0.832 {\scriptsize($\pm$0.007)}  & 3.89 {\scriptsize($\pm$0.17)} &  1.25 {\scriptsize($\pm$0.14)}   \\
  & 16 &  0 & 0.838 {\scriptsize($\pm$0.044)}  & 0.745 {\scriptsize($\pm$0.051)}  & 11.36 {\scriptsize($\pm$1.89)} & 3.30 {\scriptsize($\pm$0.78)} & 16 &   0  & 0.834 {\scriptsize($\pm$0.044)}  & 0.720 {\scriptsize($\pm$0.049)} & 8.97 {\scriptsize($\pm$1.27)} & 2.48 {\scriptsize($\pm$0.64)} \\
 & 4 &  0 & 0.710 {\scriptsize($\pm$0.051)} & 0.578 {\scriptsize($\pm$0.065)} & 19.56 {\scriptsize($\pm$1.95)} & 6.11 {\scriptsize($\pm$0.92)}  & 8  &   0  & 0.801  {\scriptsize($\pm$0.057)}  & 0.676 {\scriptsize($\pm$0.051)} & 11.40 {\scriptsize($\pm$1.53)} & 3.27 {\scriptsize($\pm$0.71)} \\ 
\hline
Volume-based LRL    &  16   &  144  & 0.892 {\scriptsize($\pm$0.016)} & 0.823 {\scriptsize($\pm$0.021)} & 7.47 {\scriptsize($\pm$0.39)} & 1.88 {\scriptsize($\pm$0.27)} &  16   &  64  &  0.871 {\scriptsize($\pm$0.010)} & 0.784 {\scriptsize($\pm$0.017)} & 5.28 {\scriptsize($\pm$0.37)} & 1.89 {\scriptsize($\pm$0.16)} \\
\cline{1-1}
IC-LRL    &  16   &  144  & 0.900 {\scriptsize($\pm$0.011)} & 0.828 {\scriptsize($\pm$0.012)} & 6.99 {\scriptsize($\pm$0.26)} & 1.75 {\scriptsize($\pm$0.22)} &  16   &  64  &  0.882 {\scriptsize($\pm$0.014)} & 0.796 {\scriptsize($\pm$0.026)} & 4.69 {\scriptsize($\pm$0.22)} & 1.66 {\scriptsize($\pm$0.12)} \\
\cline{1-1}
 GBDL (LRL) &   16   &  144  &  {\bf 0.911} {\scriptsize($\pm$0.010)}     &  {\bf 0.840} {\scriptsize($\pm$0.013)} & {\bf 6.38} {\scriptsize($\pm$0.37)} & {\bf 1.51} {\scriptsize($\pm$0.22)} &  16   &  64  &    {\bf 0.894} {\scriptsize($\pm$0.008)}  & {\bf 0.822} {\scriptsize($\pm$0.011)} & {\bf 4.03} {\scriptsize($\pm$0.41)} & {\bf 1.48} {\scriptsize($\pm$0.14)} \\
\hline
Volume-based LRL    &  4   &  156  &   0.883 {\scriptsize($\pm$0.011)}    &  0.810 {\scriptsize($\pm$0.014)}  & 8.32 {\scriptsize($\pm$0.42)} &  1.99 {\scriptsize($\pm$0.24)} &  8   &  72 & 0.865 {\scriptsize($\pm$0.017)} & 0.773 {\scriptsize($\pm$0.013)}& 6.81 {\scriptsize($\pm$0.33)} & 2.49 {\scriptsize($\pm$0.13)}  \\
\cline{1-1}
IC-LRL    &  4   &  156  &   0.889 {\scriptsize($\pm$0.014)}    &  0.814 {\scriptsize($\pm$0.014)}  & 8.01 {\scriptsize($\pm$0.31)} &  2.01 {\scriptsize($\pm$0.12)} &  8   &  72 & 0.871 {\scriptsize($\pm$0.009)} & 0.779 {\scriptsize($\pm$0.024)}& 5.96 {\scriptsize($\pm$0.22)} & 1.97 {\scriptsize($\pm$0.17)}  \\
\cline{1-1}
 GBDL (LRL)   &  4   &  156  &  {\bf 0.898}   {\scriptsize($\pm$0.008)}   &   {\bf 0.821} {\scriptsize($\pm$0.011)} & {\bf 6.85} {\scriptsize($\pm$0.44)} & {\bf 1.78} {\scriptsize($\pm$0.23)} &    8   &  72  &   {\bf 0.884} {\scriptsize($\pm$0.007)} & {\bf 0.792} {\scriptsize($\pm$0.012)} & {\bf 5.89} {\scriptsize($\pm$0.31)} & {\bf 1.60} {\scriptsize($\pm$0.15)}  \\
  \bottomrule[1pt]
  \end{tabular}}
 \vspace{-1.0ex}
 \caption{Ablation studies of different LRL variants on the KiTS19 dataset and the Atrial Segmentation Challenge dataset. For each of our results, the ``mean {\scriptsize($\pm$std)}" is reported. The first row shows the upper bound performance, i.e., all training data with their labels are used.} \vspace{-0.5ex}
 \label{tab:abla_kits_LA_1}
 \end{table*}

\begin{table*}
\centering
\resizebox{0.95\textwidth}{!}{
\begin{tabular}{@{}c|cc|cccc|cc|cccc@{}}
 \toprule[1pt]
 \multirow{3}{*}{} &  \multicolumn{6}{c|}{KiTS19} & \multicolumn{6}{c}{Atrial Segmentation Challenge} \\
  \cline{2-13}
  &  \multicolumn{2}{c|}{Scans used} &\multicolumn{4}{c|}{Metrics} & \multicolumn{2}{c|}{Scans used} &\multicolumn{4}{c}{Metrics}\\
 \cline{2-13}
 & Labeled & Unlabeled & Dice $\uparrow$ & Jaccard  $\uparrow$ & 95HD $\downarrow$ & ASD $\downarrow$ & Labeled & Unlabeled & Dice $\uparrow$ & Jaccard  $\uparrow$ & 95HD $\downarrow$ & ASD $\downarrow$ \\
 \hline
w/o $L_{MSE}$     &  16   & 144 &  0.877 {\scriptsize($\pm$0.014)}  & 0.789 {\scriptsize($\pm$0.019)} & 8.89 {\scriptsize($\pm$0.33)} &  1.96 {\scriptsize($\pm$0.19)}   &  16   &  64 &  0.873 {\scriptsize($\pm$0.012)}   & 0.781 {\scriptsize($\pm$0.011)}   & 4.76 {\scriptsize($\pm$0.46)}  & 1.66 {\scriptsize($\pm$0.17)}    \\
\cline{1-1}
w/o $L_{KL}$    &  16   &  144 &  0.868  {\scriptsize($\pm$0.013)}   & 0.776 {\scriptsize($\pm$0.009)}   & 8.99 {\scriptsize($\pm$0.29)}  & 2.03 {\scriptsize($\pm$0.17)}  &  16  &  64  &  0.876 {\scriptsize($\pm$0.014)}   & 0.775 {\scriptsize($\pm$0.016)}  & 5.12 {\scriptsize($\pm$0.55)}  & 1.87 {\scriptsize($\pm$0.15)}  \\
\cline{1-1}
Shared Encoder &   16   &  144  &  0.854 {\scriptsize($\pm$0.019)}   & 0.742 {\scriptsize($\pm$0.021)}   & 10.04 {\scriptsize($\pm$0.67)}  & 2.56 {\scriptsize($\pm$0.27)}  &  16  &  64  &  0.862 {\scriptsize($\pm$0.021)}    & 0.749 {\scriptsize($\pm$0.025)}  & 6.33 {\scriptsize($\pm$0.73)}  & 2.21 {\scriptsize($\pm$0.31)}  \\
\cline{1-1}
GBDL   &  16   &  144  &  {\bf 0.911} {\scriptsize($\pm$0.010)}     &  {\bf 0.840} {\scriptsize($\pm$0.013)} & {\bf 6.38} {\scriptsize($\pm$0.37)} & {\bf 1.51} {\scriptsize($\pm$0.22)} &  16   &  64  &    {\bf 0.894} {\scriptsize($\pm$0.008)}  & {\bf 0.822} {\scriptsize($\pm$0.011)} & {\bf 4.03} {\scriptsize($\pm$0.41)} & {\bf 1.48} {\scriptsize($\pm$0.14)} \\
\hline
w/o $L_{MSE}$     &  4   &  156  &  0.864 {\scriptsize($\pm$0.016)}   & 0.769 {\scriptsize($\pm$0.014)}   & 9.03 {\scriptsize($\pm$0.39)}  & 2.05 {\scriptsize($\pm$0.18)}  &  8   &  72 & 0.867 {\scriptsize($\pm$0.011)}   & 0.771 {\scriptsize($\pm$0.017)}  & 4.96 {\scriptsize($\pm$0.34)} & 1.85 {\scriptsize($\pm$0.20)} \\
\cline{1-1}
w/o $L_{KL}$    &  4   &  156  &   0.860 {\scriptsize($\pm$0.017)}   & 0.769 {\scriptsize($\pm$0.019)}   & 9.32 {\scriptsize($\pm$0.46)}  & 2.23  {\scriptsize($\pm$0.31)}   &  8   &  72  &  0.854 {\scriptsize($\pm$0.015)}  & 0.762 {\scriptsize($\pm$0.018)}  & 5.33 {\scriptsize($\pm$0.42)}  & 2.12 {\scriptsize($\pm$0.33)}  \\
\cline{1-1}
Shared Encoder &  4   &  156   &  0.843 {\scriptsize($\pm$0.021)}   & 0.734 {\scriptsize($\pm$0.024)}  & 11.08 {\scriptsize($\pm$0.67)}  & 3.46 {\scriptsize($\pm$0.44)}   &  8   &  72  &  0.831 {\scriptsize($\pm$0.019)}   & 0.724 {\scriptsize($\pm$0.012)}  & 7.19 {\scriptsize($\pm$0.68)} & 2.86 {\scriptsize($\pm$0.41)} \\
\cline{1-1}
GBDL   &  4   &  156  &  {\bf 0.898}   {\scriptsize($\pm$0.008)}   &   {\bf 0.821} {\scriptsize($\pm$0.011)} & {\bf 6.85} {\scriptsize($\pm$0.44)} & {\bf 1.78} {\scriptsize($\pm$0.23)} &  8   &  72  &   {\bf 0.884} {\scriptsize($\pm$0.007)} & {\bf 0.792} {\scriptsize($\pm$0.012)} & {\bf 5.89} {\scriptsize($\pm$0.31)} & {\bf 1.60} {\scriptsize($\pm$0.15)}  \\
  \bottomrule[1pt]
  \end{tabular}}
 \vspace{-1.0ex}
 \caption{Ablation studies of different parts in GBDL on the KiTS19 dataset and the Atrial Segmentation Challenge dataset. For each of our results, the ``mean {\scriptsize($\pm$std)}" is reported.}  
 \label{tab:abla_kits_LA_2}
 \vspace{-3ex}
 \end{table*}

\section{Experiments}
\label{sec:experiment}
We now report on experiments of the proposed GBDL on three public medical benchmarks. To save space, implementation details are shown in the supplementary material.  

\subsection{Datasets}
\label{sec:dataset}

The \textit{KiTS19} dataset\footnote{\url{https://kits19.grand-challenge.org/}} 
is a kidney tumor segmentation dataset, which has 210 labeled 3D computed tomography (CT) scans for training and validation. We followed the settings in previous works \cite{wang2020double, fang2020dmnet} to use 160 CT scans for training and 50 CT scans for testing. The 3D scans centering at the kidney region were used, and we used the soft tissue CT window range of [-100, 250] HU for the scans.

The \textit{Atrial Segmentation Challenge} dataset \cite{xiong2020global} includes 100 3D gadolinium-enhanced magnetic resonance imaging scans (GE-MRIs) with labels. We followed previous works \cite{wang2020double, xiong2020global, luo2021semi, huang20213d, li2020shape} to use 80 samples for training and the other 20 samples for testing.

The \textit{Liver Segmentation} dataset has been released by the MICCAI 2018 medical segmentation decathlon challenge.\footnote{\url{http://medicaldecathlon.com/index.html}} 
The dataset has 131 training and 70 testing points, and the ground-truth labels of the testing data are not released. Therefore, we used those 131 CTs in our experiments, in which 100 CT scans and 31 CT scans were used for training and testing respectively. The 3D scans centering at liver regions were utilized, and we used the soft tissue CT window range of [-100, 250] HU for the scans.

\subsection{Evaluation Metrics}
Four evaluation metrics are used as our evaluation indicators, namely, \textit{Dice Score}, \textit{Jaccard Score}, \textit{95$\%$ Hausdorff Distance (95HD)}, and \textit{Average SurfaceDistance~(ASD)}. Dice Score and Jaccard Score mainly compute the percentage of overlap between two object regions. ASD computes the average distance between the boundaries of two object regions, while 95HD measures the closest point distance between two object regions.

\subsection{Ablation Studies}
\label{sec:ablation study}

We conducted  ablation studies on the KiTS19  and the Atrial Segmentation Challenge dataset.  We comprehensively searched the hyperparameters ($\lambda_1$, $\lambda_2$, $\lambda_3$, $\lambda_4$, $\beta_1$, and $\beta_2$), and the number of sampled latent representations ($M$)  for each feedforward pass. 
Due to the space limit, the search process is shown in the
supplementary material. In all the following experiments, we fixed $\lambda_1$, $\lambda_2$, $\lambda_3$, $\lambda_4$, $\beta_1$, $\beta_2$, and $M$ to 1.0, 2.0, 1.0, 0.005, 1.0, 2.0, and 5 respectively.

Firstly, we evaluated the baseline model, i.e., the regular 3D-UNet with MC dropout that is trained with a limited number of labeled data, and we also evaluated the upper bound of the performance by utilizing all data to train the model. As Table~\ref{tab:abla_kits_LA_1} shows, when the number of labeled data is decreased, the performance of the regular 3D-UNet drops drastically.

\begin{table}
\centering 
\resizebox{0.48\textwidth}{!}{
\begin{tabular}{@{}ccccccc@{}}
 \toprule[1pt]
 \multirow{2}{*}{}    & \multicolumn{2}{c}{Scans used} &\multicolumn{4}{c}{Metrics} \\
 \cline{2-7}
 & Labeled & Unlabeled & Dice $\uparrow$ & Jaccard  $\uparrow$ & 95HD $\downarrow$ & ASD $\downarrow$ \\ 
 \hline  
 UA-MT \cite{yu2019uncertainty}     &  16  &  144 &     0.883 & 0.802 & 9.46 & 2.89  \\ 
 SASSNet \cite{li2020shape}  &  16   & 144 &  0.891 & 0.822 & 7.54 & 2.41 \\
 Double-UA  \cite{wang2020double}    &  16   &  144  &   0.895    & 0.828 & 7.42 & 2.16  \\
 Tripled-UA  \cite{wang2021tripled} &  16  &  144 &   0.887 &  0.815  & 7.55  &  2.12  \\
 CoraNet  \cite{shi2021inconsistency} & 16 &  144 &   0.898  & 0.820 & 7.23 &  1.89 \\  
 UA-MT* \cite{yu2019uncertainty}     &  16  &  144 &     0.878   & 0.797  & 8.93 & 2.76  \\
 Double-UA*  \cite{wang2020double}    &  16   &  144  &   0.899   & 0.823   & 7.55   & 2.21   \\
 Tripled-UA* \cite{wang2021tripled} &  16  &  144 &   0.886   &     0.819  & 7.43   &  2.26  \\
 CoraNet* \cite{shi2021inconsistency} & 16 &  144 &   0.894  &     0.820   & 7.37&  1.92 \\ 
 GBDL  &  16   &  144  &  {\bf 0.911}  &  {\bf 0.840} & {\bf 6.38} & {\bf 1.51} \\
 \hline  
 UA-MT \cite{yu2019uncertainty}    &  4   &  156  &    0.871      &   0.787  & 11.74 & 3.56 \\
 SASSNet \cite{li2020shape}  &  4   &  156 &  0.888 & 0.816 & 8.32 & 2.44 \\
 Double-UA  \cite{wang2020double}   &  4   &  156  &   0.887  &   0.817   & 8.04   &  2.34 \\
 Tripled-UA \cite{wang2021tripled} &  4   &  156  &   0.878 &  0.813  & 7.94  &  2.42 \\
 CoraNet \cite{shi2021inconsistency} & 4 &  156 &   0.882  &     0.814  & 8.21   &  2.44  \\  
 Double-UA*  \cite{wang2020double}   &  4   &  156  &   0.890  &     0.819   & 7.93    &  2.33  \\
 UA-MT* \cite{yu2019uncertainty}     &  4   &  156  &   0.874   &   0.790   & 11.33 & 3.21 \\
 Tripled-UA* \cite{wang2021tripled} &  4   &  156  &   0.882   &     0.810   & 7.81    &  2.47 \\
 CoraNet* \cite{shi2021inconsistency} & 4 &  156 &   0.886  &     0.812   & 8.43  &  2.39 \\
  GBDL     &  4   &  156  &  {\bf 0.898}   &   {\bf 0.821}  & {\bf 6.85}& {\bf 1.78} \\
  \bottomrule[1pt]
  \end{tabular}}
 \caption{Comparison with state-of-the-art semi-supervised segmentation methods on the KiTS19 dataset. ``*'' denotes previous methods based on the regular 3D-UNet with MC dropout.} 
 \label{tab:sota_kits}
 \vspace*{-2ex}
 \end{table}

\begin{table}
\centering 
\resizebox{0.48\textwidth}{!}{
\begin{tabular}{@{}ccccccc@{}}
 \toprule[1pt]
 \multirow{2}{*}{}    & \multicolumn{2}{c}{Scans used} &\multicolumn{4}{c}{Metrics} \\
 \cline{2-7}
 & Labeled & Unlabeled & Dice $\uparrow$ & Jaccard  $\uparrow$ & 95HD $\downarrow$ & ASD $\downarrow$ \\
 \hline  
 UA-MT \cite{yu2019uncertainty}     &  16   &  64 &    0.889      &   0.802 & 7.32 &  2.26  \\
 SASSNet \cite{li2020shape}  &  16   &  64 &  0.895 & 0.812 & 8.24 & 2.20 \\
 Double-UA  \cite{wang2020double}    &  16   &  64  &   0.897    &   0.814 &  7.04 & 2.03 \\ 
 Tripled-UA \cite{wang2021tripled} & 16 &  64 &   0.893 &     0.810  & 7.42  &  2.21  \\
 CoraNet \cite{shi2021inconsistency} & 16 &  64 &   0.887 &     0.811   & 7.55   &  2.45 \\
 Reciprocal Learning \cite{zeng2021reciprocal} & 16 & 64 & {\bf 0.901} & 0.820 & 6.70 & 2.13 \\
 DTC \cite{luo2021semi} & 16 &  64 &   0.894 &     0.810   & 7.32   &  2.10 \\ 
 3D Graph-S$^2$ Net \cite{huang20213d} & 16  & 64  & 0.898 & 0.817  & 6.68 & 2.12 \\
 LG-ER-MT \cite{hang2020local} & 16 & 64 & 0.896 & 0.813 & 7.16 & 2.06 \\
 Double-UA*  \cite{wang2020double}    &  16   &  64  &  0.894   &   0.809  &  6.16 & 2.28 \\
 UA-MT* \cite{yu2019uncertainty}     &  16   &  64 &  0.891   &   0.793   & 6.44  &  2.39 \\
 Tripled-UA* \cite{wang2021tripled} & 16 &  64 &   0.889 &     0.809  & 6.88  &  2.48  \\
 CoraNet* \cite{shi2021inconsistency} & 16 &  64 &   0.883 &     0.805  & 6.73   &  2.67 \\
 GBDL &  16   &  64  &    0.894  & {\bf 0.822} & {\bf 4.03} & {\bf 1.48}   \\
 \hline 
 UA-MT \cite{yu2019uncertainty}    &  8   &  72  &    0.843      &   0.735 &  13.83 & 3.36    \\
 SASSNet \cite{li2020shape}  &  8   &  72 &  0.873 & 0.777 & 9.62 & 2.55 \\ 
 Double-UA  \cite{wang2020double}   &  8   &  72  &   0.859      &   0.758 & 12.67 &  3.31  \\
 Tripled-UA \cite{wang2021tripled} &  8  &  72 &   0.868  &     0.768   & 10.42   &  2.98  \\
 CoraNet \cite{shi2021inconsistency} & 8 &  72 &   0.866  &     0.781   & 12.11  &  2.40   \\
 Reciprocal Learning \cite{zeng2021reciprocal} & 8 & 72 & 0.862 & 0.760 & 11.23 & 2.66 \\
 DTC \cite{luo2021semi} & 8 &  72 &   0.875 &     0.782   & 8.23   &  2.36 \\ 
 LG-ER-MT \cite{hang2020local} & 8 & 72 & 0.855 & 0.751 & 13.29 & 3.77 \\
 3D Graph-S$^2$ Net \cite{huang20213d} & 8  & 72  & 0.879 & 0.789  & 8.99 & 2.32\\
 Double-UA*  \cite{wang2020double}   &  8   &  72  &   0.864  &   0.767  & 10.99 &  3.02  \\
 UA-MT* \cite{yu2019uncertainty}      &  8   &  72 &    0.847    &   0.744   & 12.32  &  3.20 \\
 Tripled-UA* \cite{wang2021tripled} &  8  &  72 &   0.868 &     0.760  & 9.73  &  3.31 \\
 CoraNet* \cite{shi2021inconsistency} & 8 &  72 &   0.861 &     0.770 & 11.32    &  2.46  \\
 GBDL  &  8   &  72  &   {\bf 0.884} & {\bf 0.792}& {\bf 5.89}  & {\bf 1.60}  \\
  \bottomrule[1pt]
  \end{tabular}}
 \caption{Comparison with state-of-the-art semi-supervised segmentation methods on the Atrial Segmentation Challenge. ``*'' denotes previous methods based on the regular 3D-UNet with MC dropout.} 
 \label{tab:sota_LA}
 \end{table}


Second, as for an input volume, LRL assumes that the latent representation of each slice follows a multivariate Gaussian distribution, and therefore, the latent representation of the whole volume is sampled from a joint Gaussian distribution that is obtained by multiplying the Gaussian distribution PDFs of those slices. 
However, LRL can also be designed to learn the latent representation of the whole volume, i.e., estimating the joint Gaussian distribution directly, and we denote this variant as ``volume-based LRL''.  Compared to the original LRL, this variant is much closer to a 3D-CNN-based cVAE, since each input volume is treated as a whole, and the volume-based LRL represents the whole volume with a latent representation, similar to the original cVAE that represents the whole input image with a latent representation. 
In this case, 
$L_{KL[Q(Z|X)||P(Z)]}$ degenerates to $1/2\cdot(-log\, det [\Lambda_v^{-1}] +  tr[\Lambda_v^{-1}] +   \mu_v ^T \mu_v - {D})$, where $\mu_v$ and $\Lambda_v$ are the mean vector and the precision matrix of the joint Gaussian distribution, respectively. 
We evaluated such volume-based LRL, and the results are displayed in Table~\ref{tab:abla_kits_LA_1}. For a fair comparison, the $\mu_v$ and $\Lambda_v$ are of the same number of dimensions as the $\mu_i$ and $\Lambda_i$ in the original LRL. The results show that the volume-based LRL performs worse than the original LRL that is presented in Section~\ref{sec:gbdl}, demonstrating that the original LRL can better model the joint distribution. Based on this observation, we hypothesize that keeping a difference among slices is important, since slices and their corresponding segmentation masks are different from each other. The original LRL adds perturbation (random noise) to each slice, which is beneficial to keeping such a difference. However, the volume-based LRL only adds random noise to the latent representation of the whole volume, which goes against to keeping such a difference.

Third, considering that the independent components assumption is widely
used in cVAE, we also evaluated LRL variant with this assumption, and we denote this variant as ``IC-LRL''.  
In this case,  $L_{KL[Q(Z|X)||P(Z)]}$ degenerates to $1/2 \cdot \sum_{D}[-log\frac{1}{\sum_{i=0}^{n-1}\frac{1}{\sigma_i^2}}
+ \frac{1}{\sum_{i=0}^{n-1}\frac{1}{\sigma_i^2}} + (\frac{1}{\sum_{i=0}^{n-1}\frac{1}{\sigma_i^2}}\sum_{i=0}^{n-1}\frac{\mu_i}{\sigma_i^2})^2  - 1 ]$, where $\mu_i$ and $\sigma_i$ are the mean vector and the variance vector of the Gaussian distribution for each slice in the input volume, respectively. The results shown in Table~\ref{tab:abla_kits_LA_1} indicate that the IC-LRL performs worse than our original LRL, verifying that the independent components assumption harms the performance of LRL.

Furthermore, we analyzed the two important parts of the proposed LRL, i.e., the reconstruction loss $L_{MSE}$ and the term $L_{KL[Q(Z|X)||P(Z))]}$. We removed these two loss terms,\footnote{Removing $L_{MSE}$ is equivalent to removing the reconstruction path, i.e., when we did an ablation study about removing the reconstruction loss $L_{MSE}$, we also removed the upper 3D-UNet decoder of LRL in Figure~\ref{eqn:overall}.} respectively, to study their impact on GBDL. As Table~\ref{tab:abla_kits_LA_2} shows, when $L_{MSE}$ or $L_{KL[Q(Z|X)||P(Z))]}$ is removed, a performance decline can be observed, demonstrating the importance of these two terms and the correctness of our Bayesian formulation.

 \begin{table}
\centering
\resizebox{0.45\textwidth}{!}{
\begin{tabular}{@{}ccccccc@{}}
 \toprule[1pt]
 \multirow{2}{*}{}    & \multicolumn{2}{c}{Scans used} &\multicolumn{4}{c}{Metrics} \\
 \cline{2-7}
 & Labeled & Unlabeled & Dice $\uparrow$ & Jaccard  $\uparrow$  & 95HD $\downarrow$ & ASD $\downarrow$  \\
 \hline 
 3D-UNet w/ MC dropout &   100   &  0    & 0.950   & 0.899   & 6.04   & 1.53  \\ 
 \hline 
 UA-MT \cite{yu2019uncertainty}     &   5   &  95   &    0.920   & 0.867   & 13.21    & 4.54    \\ 
 Double-UA  \cite{wang2020double}    &   5   &  95   &   0.927   &   0.878   & 12.11   & 4.19   \\
 Tripled-UA  \cite{wang2021tripled}   &   5   &  95   &   0.921  &   0.869    & 11.77   & 3.62   \\
 CoraNet \cite{shi2021inconsistency} & 5 &  95 &  0.923  &  0.877   & 10.84  & 4.28 \\ 
 GBDL    &   5   &  95   &   {\bf 0.935}  &  {\bf 0.884} & {\bf 7.89}  & {\bf 2.42}  \\
 \bottomrule[1pt]
\end{tabular}}
\vspace{-1ex}
 \caption{Comparison with state-of-the-art semi-supervised segmentation methods on the Liver Segmentation dataset. Previous methods in this table are also based on the regular 3D-UNet with MC dropout.}  
\label{tab:sota_liver}

\end{table}

\begin{table}
\centering 
\resizebox{0.4\textwidth}{!}{
\begin{tabular}{@{}c|c|c|c|c@{}}
 \toprule[1pt]
 \multirow{2}{*}{Method} & \multicolumn{2}{c|}{KiTS19}  & \multicolumn{2}{c }{Atrial} \\
 \cline{2-5}
       & 16 labeleld  & 4 labeleld & 16 labeleld  & 8 labeleld \\
 \hline  
 UA-MT \cite{yu2019uncertainty}  &  0.868  & 0.841 &  0.851  & 0.836 \\
 Double-UA \cite{wang2020double}  &  0.881  &  0.864 &  0.861  & 0.842 \\
 Triplet-UA  \cite{wang2021tripled} &  0.874  &  0.852 &  0.864  & 0.847  \\
 CoraNet  \cite{shi2021inconsistency} &  0.882  &  0.866 &  0.867  & 0.855 \\
 GBDL  &  {\bf 0.897}  &  {\bf 0.877} &  {\bf 0.882}  & {\bf 0.864} \\ 
  \bottomrule[1pt]
  \end{tabular}} 
 \caption{  PAvPU of different Bayesian deep learning methods.} 
 \label{tab:pavpu} 
 \end{table}

\begin{figure*}[t] 
\centering
\includegraphics[width=0.75\linewidth]{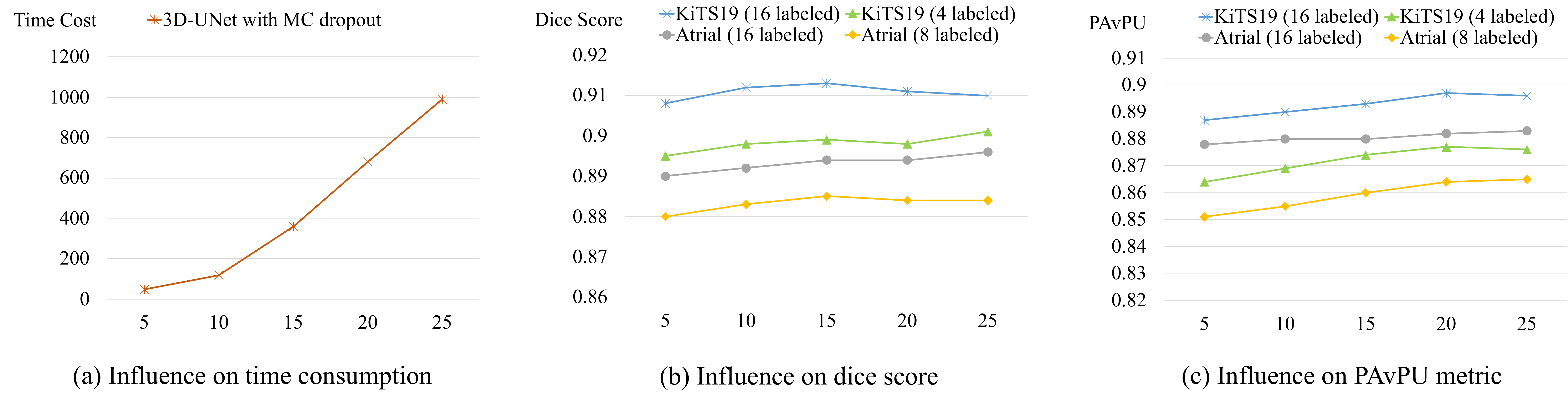} 
 
\caption{The influence of the number of feedforward passes on: (a) time costs (ms) for processing a $128\times128\times32$ volume input on a single GeForce GTX 1080 Ti; (b) Dice score; and (c) PAvPU metric.  The horizontal axis refers to the number of feedforward passes, namely, $T$.}
\vspace{-1ex}
\label{fig:runtime}
\end{figure*} 

Besides, as the two 3D-UNet encoders in  LRL are of the same architecture (see the supplementary material), and both aim at extracting features from input volumes, we evaluated a case where the two encoders share their weights. Nevertheless, sharing the weights degrades the performance, as shown in Table~\ref{tab:abla_kits_LA_2}. Similarly, although some layers of the two 3D-UNet decoders in the LRL also have the same configuration, sharing their weights  would  clearly lead to a lower performance, as the respective objectives of them are different, leading to different gradient descent directions.

Finally, when compared with the baseline model, the proposed GBDL can improve the performance with a huge margin with respect to the four assessment metrics for both datasets. So, the proposed method is a good solution for  semi-supervised volumetric medical imaging segmentation.

 \subsection{Comparison with State-of-the-Art Methods}
The proposed GBDL is compared with previous state-of-the-art methods, and since previous works provide no standard deviation, we also only report the mean values of the four evaluation metrics in Tables~\ref{tab:sota_kits}, \ref{tab:sota_LA}, and \ref{tab:sota_liver}. 
Since most previous methods are based on VNet, for fair comparison, our regular 3D-UNet with MC dropout is designed to have similar numbers of parameters. Furthermore, we also re-implement previous Bayesian deep learning methods based on the regular 3D-UNet with MC dropout, and according to the results in Tables~\ref{tab:sota_kits} and \ref{tab:sota_LA},  their performance are similar to their VNet counterparts, meaning that the comparison between our method and previous methods is relatively fair.

By  Tables~\ref{tab:sota_kits},  \ref{tab:sota_LA}, and \ref{tab:sota_liver}, the performance of GBDL is higher than previous state-of-the-art results on the three public medical datasets. When the number of labeled data decreases, the performance gap between our GBDL and previous state-of-the-art methods becomes larger, which indicates that our method is more robust when fewer labeled data were used for training. More importantly, GBDL performs better than all previous Bayesian deep-learning-based methods \cite{yu2019uncertainty, wang2020double, shi2021inconsistency, wang2021tripled}, which verifies that GBDL is better than the teacher-student architecture used in them and that the generative learning paradigm is more suitable for pseudo-label generation and solving the semi-supervised segmentation task than the discriminative counterpart.

Furthermore, we compare the patch accuracy vs. patch uncertainty (PAvPU) metric \cite{mukhoti2018evaluating} for GBDL
and previous Bayesian-based methods to evaluate their uncertainty estimation. By   Table~\ref{tab:pavpu}, GBDL can output more reliable and well-calibrated uncertainty estimates than other methods, so that clinicians can do a post-processing and refine the results in practice based on the uncertainty given by GBDL.

Finally, since full Bayesian inference is time-consuming, we show the relationship among the number of feedforward passes $T$, the time consumption, and the performance in Figure~\ref{fig:runtime}. 
By Figure~\ref{fig:runtime}(a), 
the time costs rise with increasing~$T$, but by Figure~\ref{fig:runtime}(b) and (c), the Dice score and the PAvPU metric are not greatly impacted by $T$. Thus, $T=5$ can be chosen for saving time to get the uncertainty in practice, with only a minor performance degradation.

\section{Summary and Outlook}
\label{sec:conclusion}

In this paper, we rethink the issues in previous Bayesian deep-learning-based methods for  semi-supervised volumetric medical image segmentation, and have designed a new generative Bayesian deep learning (GBDL) architecture to solve them. The proposed method outperforms previous state-of-the-art methods on three public medical benchmarks, showing its effectiveness for handling the data with only a very limited number of annotations. 

However, in GBDL, there is still a limitation with regard to the time consumption for full Bayesian inference and  uncertainty estimation.  By Figure~\ref{fig:runtime}(a), several numbers of feedforward passes will lead to a high cost in terms of time. Since Bayesian deep learning is important for some safety-critical areas, such as medical diagnosis and autonomous driving, future research can try to find  better methods to improve the inference speed for GBDL. Moreover, since GBDL is a new proposed architecture, future research can also consider to combine other previous deep-lear\-ning-based methods (such as different training and learning strategies) with GBDL to further improve the performance.

\paragraph{Acknowledgments. }%
This work was partially supported by the Alan Turing Institute under the EPSRC grant EP/N510129/1, by the AXA Research Fund, by the EPSRC grant EP/R013667/1, and by the EU TAILOR grant. We also acknowledge the use of the EPSRC-funded Tier 2 facility JADE (EP/P020275/1) and GPU computing support by Scan Computers International Ltd.

\newpage
{\small
\bibliographystyle{ieee_fullname}
\bibliography{egbib}
}

\end{document}